\documentclass{article} % For LaTeX2e
\usepackage{iclr2027,times}

\usepackage{amsmath,amsfonts,bm}

\def\eqref#1{equation~\ref{#1}}
\def\1{\bm{1}}

\DeclareMathAlphabet{\mathsfit}{\encodingdefault}{\sfdefault}{m}{sl}
\SetMathAlphabet{\mathsfit}{bold}{\encodingdefault}{\sfdefault}{bx}{n}

\usepackage{hyperref}
\usepackage{url}
\usepackage{enumitem}

\usepackage{graphicx}
\usepackage[misc]{ifsym}
\usepackage{amsmath}
\usepackage{mathtools}
\usepackage[capitalize,noabbrev]{cleveref}
\usepackage{booktabs}
\usepackage{multirow}
\usepackage{pifont}
\usepackage{algorithm}
\usepackage{algpseudocode}
\usepackage[most]{tcolorbox}
\usepackage[dvipsnames]{xcolor}
\title{Beyond Context Length: Characterising the Difficulty of Long-Context Reasoning}

\author{%
  Shiwen Qin, Cameron Barker, Henry Gouk, Elliot J.~Crowley,\\ \textbf{Shay B.~Cohen}, \textbf{Mirella Lapata}, \textbf{Miao Li}\thanks{Miao Li is the corresponding author and supervised Shiwen Qin in conducting this research.}\\
  University of Edinburgh, Edinburgh, United Kingdom\\
  \{shiwen.qin, cameron.barker, henry.gouk, elliot.j.crowley\}@ed.ac.uk, \\\{scohen@inf., mlap@inf., miao.li@\}ed.ac.uk
}

\iclrfinalcopy % Uncomment for camera-ready version, but NOT for submission.
\begin{document}

\maketitle

\begin{abstract}

Long-context language models support ever longer context windows, and evaluations typically treat context length as the main indicator of difficulty. However, context length alone cannot explain why models fail on long-context tasks that require reasoning over complex information. Contexts of equal length can pose very different challenges: a task may be solvable from a single local passage, or it may require integrating evidence distributed across the full context. We quantitatively characterise this difficulty through three properties: \textit{evidence scope}, how much necessary information must be found; \textit{evidence dispersion}, how widely that information is scattered within the context; and \textit{reasoning depth}, how many dependent steps are needed to derive an answer from it. We construct \texttt{LURCH}, a multi-domain question-answering benchmark of 6,550 instances with supporting evidence and reasoning graphs, enabling all three properties to be quantified. Evaluations of proprietary and open-weight models show that: (1) at fixed context length, all our three properties are strongly negatively correlated with model performance; (2) as context length increases, performance degrades sharply on samples with dispersed evidence, whereas it remains relatively stable on samples with localised evidence; and (3) difficulty compounds across properties, as questions with moderately high values on all three are harder than those with an extreme value on just one. These findings establish evidence scope, dispersion and reasoning depth as measures of long-context reasoning difficulty that complement context length.
\end{abstract}

% ML: it is unlcear what scope is. Wht are these properties of? Reaasoning is a property of the model, scope is of the question and dispersion of the context?

\begin{figure}[H]
  \centering
  \includegraphics[width=0.75\linewidth]{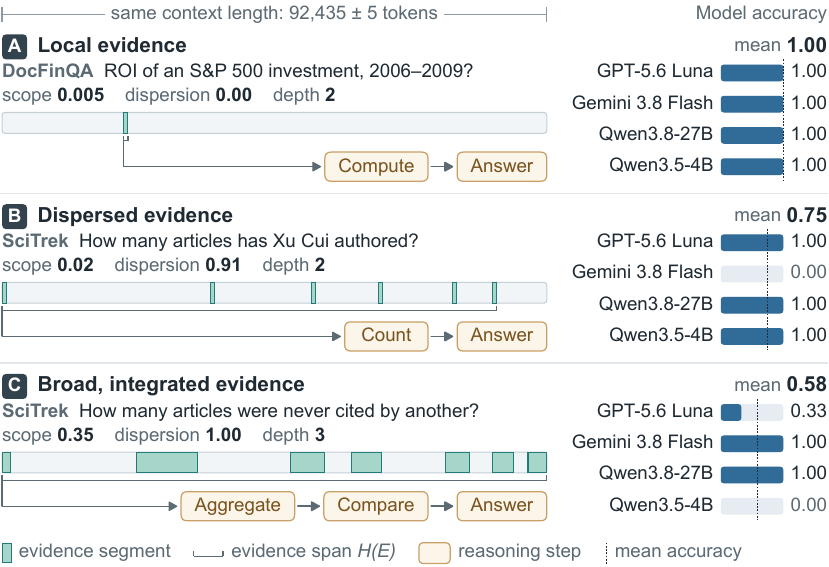}
  \vspace{-10pt}
  \caption{Equal input length,  different difficulty due to evidence scope, dispersion and reasoning depth.}
  \label{fig:teaser}
  % \vspace{-2pt}
\end{figure}

\section{Introduction}
\label{sec:introduction}

Large language models (LLMs) can now process hundreds of thousands or even millions of tokens enabling applications involving books, scientific papers, financial reports, and long conversation histories \citep{team2023gemini, yang2025qwen3, singh2025openai, xu2026deepseek}. However, the ability to accommodate long inputs does not guarantee effective reasoning over them: models struggle to use information located in the middle of the context \citep{liu2024lost}, and their effective context length often falls well short of the claimed window \citep{hsieh2024ruler}. Existing evaluations typically treat context length as the main indicator of difficulty, assessing models on the same tasks at different input lengths~\citep{hsieh2024ruler, lu2026corpusqa, li2026scitrek}. While this approach reveals how performance changes with length, it leaves open which properties beyond token count make long-context reasoning difficult. 
%As shown in Figure~\ref{fig:teaser}, contexts of equal length can pose very different challenges: a task may be solvable from a single local passage, or it may require integrating evidence distributed across the full context.

Long-context benchmarks have increasingly moved from testing whether models can retrieve information to testing whether they can reason over it, covering multi-document, domain-specific, and aggregation-heavy tasks \citep{loong_2024, reddy2024docfinqa, bai2025longbench, li2026scitrek}. Yet these benchmarks mainly reveal \emph{where} models fail, rather than \emph{what} makes individual instances difficult. \citet{goldman2024really} take an important step by conceptualising two dimensions of long-context difficulty: \textit{scope} (``How much necessary information is there to find?'') and \textit{dispersion} (``How hard is it to find them in the context?''). These concepts help explain why questions of similar length may impose different demands on a model, but they are defined qualitatively rather than as measurable quantities. ETHIC~\citep{lee2025ethic} quantifies information coverage, but relies on an evaluator model's judgements of which context chunks are potentially necessary, and captures only one aspect of difficulty in a limited range of task settings. Without quantitative definitions, it is difficult to establish how these properties vary across tasks, how they relate to model performance, and whether they account for differences in difficulty beyond context length.

% ML: scope, dispersion and maybe reasoning should be in abstract

% Prior work~\citep{goldman2024really} conceptualizes the difficulty of long-context tasks along two intuitive dimensions: scope (``How much necessary information is there to find?'') and dispersion (``How hard is it to find them in the context?'').
% These concepts help explain why questions with similar input lengths may impose different demands on a model. However, this conceptual definition does not provide quantitative and measurable definitions to scope and dispersion.
% Without quantitative definitions, it is difficult to establish how these properties vary across existing benchmarks, how they relate to model performance, and whether they account for differences in difficulty beyond input length.
% In this work, we introduce quantitative, sample-level measures of both properties, based on the amount and positions of supporting evidence.
% We also propose reasoning depth, the longest dependency path in an evidence-to-answer reasoning graph. Parallel evidence lookups do not increase depth; operations that depend on earlier results do. Together, these properties characterise how much supporting information is involved, where it occurs, and how it is combined to derive an answer.

In this work, we propose quantitative definitions of \textit{evidence scope} and \textit{evidence dispersion}, based on the amount and positions of the supporting evidence within the context. We further extend this characterisation with \textit{reasoning depth}, defined as the longest dependency path in an evidence-to-answer reasoning graph: parallel evidence lookups do not increase depth, whereas operations that depend on earlier results do. Together, these three properties characterise how much supporting information is involved, where it occurs, and how it must be combined to derive an answer. To study how these properties contribute to model failures, we construct \texttt{LURCH} (Long-context Understanding and Reasoning from Contextual Hints), a multi-domain question-answering benchmark built from diverse, naturally occurring long-context tasks. Each question is provided with its answer, source-provided or automatically annotated supporting evidence, and an answer-conditioned reasoning graph, enabling all three properties to be quantified. Using \texttt{LURCH}, we conduct controlled experiments on multiple LLMs to address three research questions:
\begin{enumerate}[leftmargin=*, label=\textbf{RQ\arabic*:}]
\item \textbf{Difficulty at fixed length.} Among instances of the same context length, how strongly are evidence scope, evidence dispersion, and reasoning depth associated with model performance?

\emph{Finding:} 
%At fixed context length, task, and model, all three properties are strongly negatively correlated with performance.
Across questions, greater scope, dispersion and reasoning depth are each associated with lower scores after accounting for context length, source dataset and model.
Notably, doubling reasoning depth is associated with a score drop comparable to that of doubling context length.

\item \textbf{Length-dependent difficulty.} How does the effect of context length on performance vary with evidence scope, evidence dispersion, and reasoning depth?

\emph{Finding:} As context length increases, performance degrades sharply on instances whose evidence is widely dispersed, whereas it remains relatively stable on instances with localised evidence. This pattern holds across all evaluated models.

\item \textbf{Compounding difficulty.} Are model failures associated with particular combinations of scope, dispersion, and depth, even when no individual property is extreme?

\emph{Finding:} Difficulty compounds across properties: questions with moderately high values on all three properties are harder than those with an extreme value on just one. This pattern is broadly consistent with an additive combination of the three properties.

\end{enumerate}

These findings establish evidence scope, evidence dispersion, and reasoning depth as measures of long-context reasoning difficulty that complement context length, and provide a basis for diagnosing model failures and designing more informative long-context evaluations.

\section{Related Work}
\label{sec:related_work}

\paragraph{Long-context language models.}

% Effective modelling of long language sequences has been a longstanding goal in natural language processing. A series of technological innovations has made it possible for LLMs to process increasingly long contexts.
Both closed (e.g., Gemini 3.8 Flash, Opus 5.5, and GPT-6 Astra) and open (e.g., GLM-5.3~\citep{glm5_2026}) weight models now support context windows of up to 1M tokens.
Advances in positional encoding, including Position Interpolation and YaRN \citep{chen2023extending,peng2024yarn}; efficient attention mechanisms, such as FlashAttention, ring attention, and sparse attention \citep{dao2022flashattention,liu2024ringattention,llama4_2025}; and KV-cache efficiency techniques, including GQA, PagedAttention, and MLA \citep{gqa_2023,pagedattention_2023,deepseek_v2_2024}, have made increasingly long contexts practical.
Despite these advances, extending a model's context window does not guarantee perfect use of the information it contains.
\citet{liu2024lost} find performance degradation when critical information occurs in the middle of the context.
RULER \citep{hsieh2024ruler} further shows the effective context length of most models, including both open-source and proprietary ones, rarely exceeds half of the claimed length.
These findings suggest that context length alone does not capture the difficulty that long-context inputs pose to language models. Yet, the factors that determine the failures of long-context language models remain largely under-explored.

\paragraph{Reasoning over long contexts.}

Early benchmarks primarily examined whether models could locate and retrieve relevant information from long contexts (e.g., RULER~\citep{hsieh2024ruler}, NeedleBench~\citep{needlebench_2024}).
More recent work extends long-context benchmarks to diverse reasoning-intensive tasks. LongBench v2~\citep{bai2025longbench} includes document question answering, dialogue-history understanding, and code-repository understanding grounded on human-written questions and answers, while Loong~\citep{loong_2024} focuses on reasoning questions on multiple input documents. 
Domain-specific benchmarks further target particular reasoning demands. DocFinQA \citep{reddy2024docfinqa} evaluates financial reasoning using full-document contexts, while SciTrek~\citep{li2026scitrek} requires counting, aggregation, and comparison across collections of scientific articles. Together, these benchmarks reflect a broader shift from evaluating whether models can retrieve information from long contexts to evaluating whether they can deliver complex reasoning over them.
However, existing benchmarks primarily reveal where models fail, but generally do not tell us what makes individual reasoning instances difficult. 
This limits our ability to identify the sources of difficulty in long-context reasoning and develop targeted approaches to address them. 
Recent work has begun to address this gap by examining factors that may contribute to reasoning difficulty. \citet{goldman2024really} conceptualise scope and dispersion as complementary conceptual axes of long-context tasks derived from extensive human annotation but without explicit quantification, distinguishing the amount of information required from the difficulty of locating it within the context. ETHIC~\citep{lee2025ethic} operationalises information coverage by quantifying the proportion of context chunks that an evaluator model deems potentially necessary to answer a query. Existing efforts, however, remain limited in both task coverage and conceptual scope, focusing on a relatively narrow range of reasoning settings and covering only a subset of the factors that may determine instance-level difficulty. As a result, we still lack a systematic framework for characterising the difficulty of long-context reasoning across diverse tasks.

\section{Reasoning Characterisation}
\label{sec:characterisation}

We complement context length with three properties: \textbf{evidence scope}, \textbf{evidence dispersion}, and \textbf{reasoning depth}. \cref{fig:teaser} illustrates these distinctions: examples A and B both require relatively little evidence and have reasoning depth two, but differ substantially in evidence dispersion. Example C combines broader evidence coverage with high dispersion and greater reasoning depth.

\paragraph{Notation.}
Let $L$ denote the length of the document in tokens, and let the evidence set contain $N$ sorted, non-overlapping evidence intervals $E=\{e_1,e_2,\ldots,e_N\}$ where each interval $e_i = [a_i, b_i]$ has length $\ell_i=b_i-a_i+1$. We merge overlapping or adjacent intervals, so the resulting intervals are disjoint and separated by at least one non-evidence token. We denote the total number of evidence tokens by $S(E)=\sum_{i=1}^N \ell_i.$

% ML: Possibly can be explained better using an illustrative example
\paragraph{Evidence scope.}
Scope measures the proportion of the context covered by the evidence set,
\begin{equation}
  \mathrm{Scope}(E)
  =
  \frac{S(E)}{L}.
  \label{eq:scope}
\end{equation}

% old
%Scope lies in $[0,1]$, with a small value indicating that the answer can be supported using a relatively small portion of the context, and a large value indicates that a substantial proportion of the context is required. Scope measures the amount of required information but does not describe its location. For example, two instances may have identical scope even if the evidence in one forms a compact block while the evidence in the other is distributed sparsely between the beginning and end of the document. We introduce evidence dispersion to distinguish these cases.
%Scope ($\in [0,1] $) measures the evidence fraction; dispersion distinguishes a single evidence block from scattered passages of equal scope.

\paragraph{Evidence dispersion.}
Dispersion measures the extent to which the evidence spans the document, relative to the amount of non-evidence text. We define the \textbf{evidence span} as the distance from the beginning of the first interval to the end of the last interval, $H(E)=b_N-a_1+1.$ (Note that $H(E) \ge S(E)$.) The number of non-evidence tokens lying between the first and last evidence intervals and the total number of non-evidence tokens in the context are, respectively,

\begin{equation}
  G_{\mathrm{internal}}(E)=H(E)-S(E),
  \qquad \text{and} \qquad
  G_{\mathrm{total}}(E)=L-S(E).
  \label{eq:gaps}
\end{equation}
We define evidence dispersion as the fraction of all non-evidence text that lies inside the evidence span,
\begin{equation}
  \operatorname{Dispersion}(E)=
  \dfrac{G_{\text{internal}}}{G_{\text{total}}}.
  \label{eq:dispersion}
\end{equation}

Dispersion lies in $[0,1]$: values near zero indicate clustered evidence, while values near one indicate that most non-evidence text lies between the evidence intervals. It is zero for a single interval and one when the evidence span covers the full context ($a_1=0$, $b_N=L-1$).

\paragraph{Reasoning depth.}
Reasoning depth measures the sequential dependencies involved in deriving an answer from evidence. We represent a derivation as a directed acyclic graph, in which nodes denote atomic operations, including evidence lookups and inference steps, and edges denote direct dependencies between operations. Depth is the number of nodes on the longest dependency path ending in an answer-producing operation. This representation accommodates both sequential inference and the synthesis of evidence obtained in parallel; it does not assume that reasoning forms a single chain. For example, the question ``Which film was released first, Bichunmoo or Musa?'' needs the release year of each film. The two years are retrieved independently at depth one, and comparing them is one further step at depth two (\cref{fig:reasoning-depth-example} in Appendix~\ref{appendix:reasoning_validation}). Additional parallel lookups do not increase depth unless further sequential operations are needed to produce the answer. Under this convention, an answer read directly from the evidence has depth one. Depth does not count the total number of operations or the difficulty of individual steps.

\section{Benchmark Construction}
\label{sec:benchmark_construction}

\begin{figure}[t!]
  \centering
  \includegraphics[width=0.8\linewidth, clip]{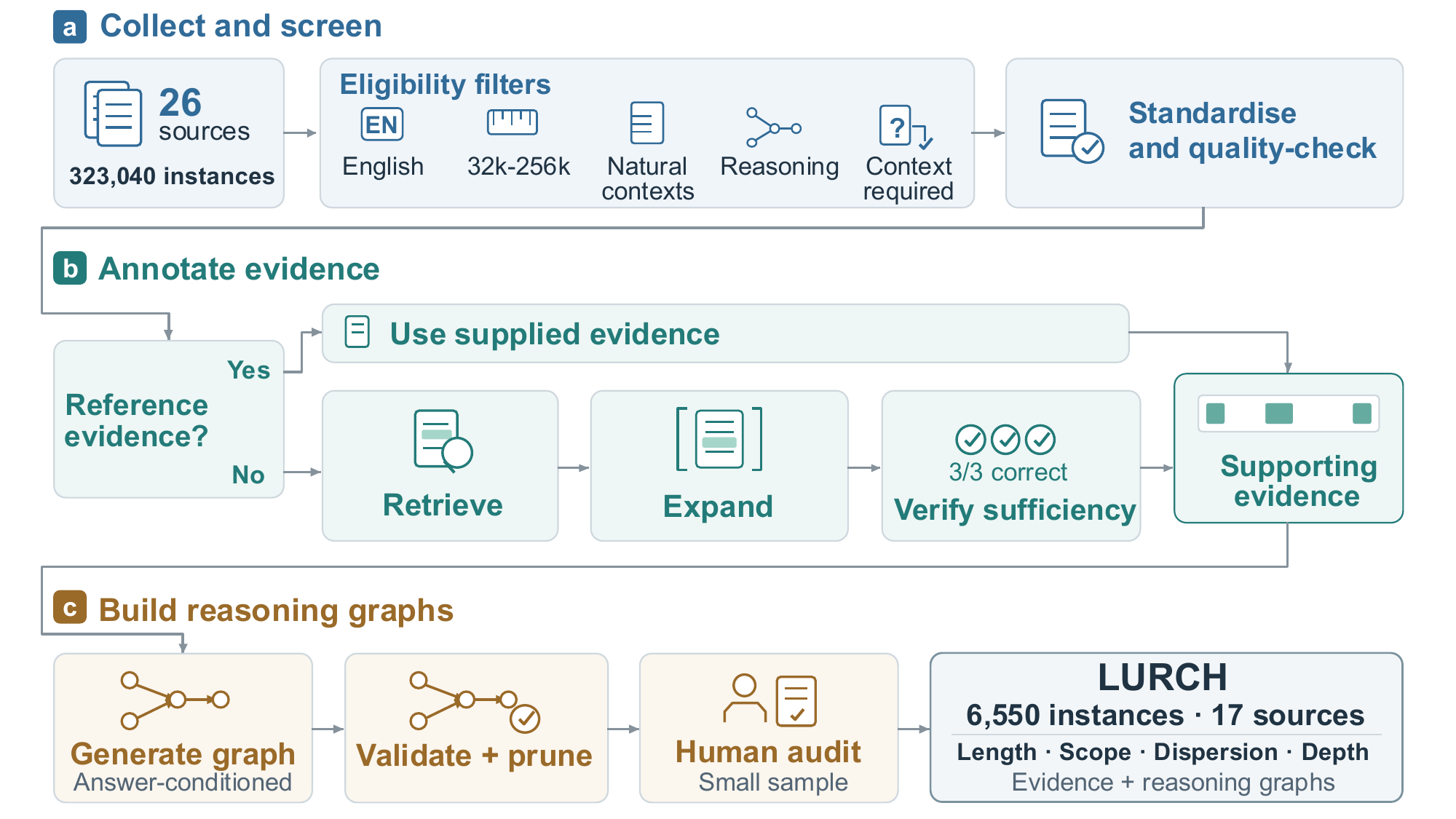}
  \vspace{-10pt}
  \caption{The dataset construction pipeline of \texttt{LURCH}.}
  \label{fig:pipeline}
  \vspace{-15pt}
\end{figure}

We construct \texttt{LURCH} by standardising instances from existing long-context benchmarks, using reference or automatically annotated evidence, and generating validated reasoning graphs (\cref{fig:pipeline}).

\subsection{Data Collection}
\label{sec:data_collection}

Filtering, validation and subsampling reduce 323,040 instances from 26 candidate sources to 6,550 instances from 17 sources. Per-source statistics are in \cref{tab:task_statistics}; Appendix~\ref{appendix:data_source} catalogues all candidate sources, including those excluded.

\paragraph{Multi-stage data filtering.}
We retain English-language instances with naturally occurring contexts of 32k--256k tokens, prioritising tasks involving multi-hop reasoning, comparison, aggregation, or evidence integration rather than single-fact retrieval or span extraction. We also exclude questions answered correctly by a screening model without the context, reducing the likelihood that they can be solved from parametric knowledge alone. Filtering details are in Appendix~\ref{appendix:data_filtering}.

\paragraph{Unified prompts and evaluation metrics.}
We preserve the original task instructions while standardising the request for step-by-step reasoning and the final-answer format across sources. We evaluate final answers using Exact Match or token-level F1, as appropriate for each task. For Loong~\citep{loong_2024}, we follow the original LLM-as-a-judge protocol to assess open-ended responses for which lexical overlap alone may not capture answer quality accurately. Per-source metrics and evaluation details are provided in Appendix~\ref{appendix:data_statistics}.

\paragraph{Question and answer quality screening.}
To identify quality issues not captured by automated filters, two authors manually inspect five randomly sampled instances from each remaining source. The review checked whether (i) reference answers were non-empty and complete, (ii) answer formats satisfied the question’s requirements, and (iii) questions were unambiguous and answerable from the supplied context. We exclude a source if any inspected instance fails these checks. This small-sample review serves as a source-level screening step rather than an exhaustive validation of the retained instances.

\subsection{Evidence Annotation}
\label{sec:evi_retrieval}

For each question, we identify supporting evidence within its source context to measure scope and dispersion and construct reasoning graphs. We use benchmark-provided reference evidence where available. Otherwise, we retrieve candidate passages, expand them to their containing paragraphs or full tables, and merge overlapping or adjacent intervals. A verifier then answers the question using only the expanded evidence, without access to the reference answer; We sample three answers from the verifier for each question and retain the question only if all three meet the task-specific correctness criterion when compared with the reference answer. This repeated check reduces reliance on a single successful response. Scope and dispersion therefore describe the expanded supporting context, not necessarily a minimal evidence set (more details in Appendix~\ref{appendix:evidence_annotation}).

\subsection{Reasoning Graph Generation}
\label{sec:reasoning_graph}

After evidence retrieval and sufficiency validation, we use two separate annotators, Gemini 3.7 Flash and GPT-5.6 Terra, to construct answer-conditioned reasoning graphs. Each receives the question, reference answer, and numbered evidence statements and reconstructs a minimal evidence-to-answer derivation. We validate each graph by removing malformed or duplicate steps, invalid references, and cycle-forming edges, then prune steps that do not contribute to the answer; unparseable or empty graphs are invalid. Across all 6,550 instances, the two annotators assign identical reasoning depths in 65.6\% of cases, with a Spearman rank correlation of 0.48 between their depth estimates. Two human auditors compare the two graphs on 40 instances. Neither auditor prefers GPT graphs more often than Gemini graphs, and one prefers Gemini on 14 of the 20 instances where the depths differ. We therefore compute the graph-derived properties using only Gemini generations. Details of annotator agreement, the audit procedure, and audit results are provided in the Appendix~\ref{appendix:reasoning_validation}. We compute reasoning depth from these graphs (\cref{sec:characterisation}).

\subsection{Dataset Properties}
\label{sec:attributes}

\begin{figure}
  \centering
  \includegraphics[width=0.8\linewidth]{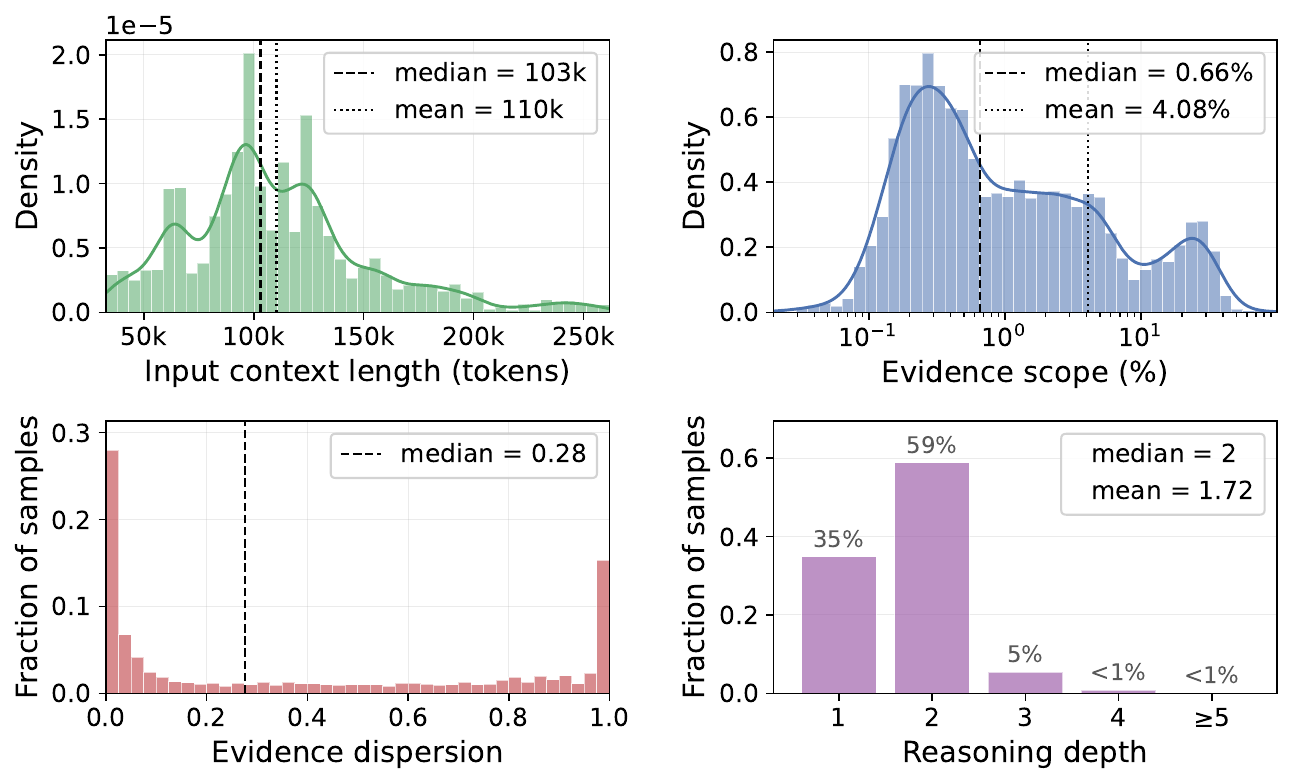}
  \vspace{-10pt}
  \caption{Distributions of input context length, evidence scope, evidence dispersion, and reasoning depth. Dashed and dotted lines indicate the median and mean, respectively; zero dispersion denotes evidence contained in a single contiguous block.}
  \label{fig:attribute_distributions}
  \vspace{-15pt}
\end{figure}

\cref{fig:attribute_distributions} shows context length and the three properties over all instances.
Contexts have a median of 103k tokens and a mean of 110k. Within them, the supporting evidence is sparse, with a median of just 0.66\% of the context and spanning more than three orders of magnitude.
Only 11\% of instances have all evidence in a single contiguous block, and the median dispersion among the rest is 0.39.
Reasoning depth is mostly multi-step: 65\% of instances require two or more dependent steps. The benchmark therefore couples long inputs with sparse, widely dispersed evidence and non-trivial compositional reasoning. 
For individual tasks, the ones with similar median context lengths exhibit different evidence profiles: HotpotQA has narrow, localised evidence, whereas SciTrek and CorpusQA have broader, highly dispersed evidence (Appendix~\ref{appendix:attr_distribution_per_task}).

\section{Experiments}
\label{sec:experiments}

\subsection{Experimental Settings}
\label{sec:exp_procedure}

% \paragraph{LLMs and data.}
We evaluate two proprietary LLMs, GPT-5.6 Luna and Gemini 3.8 Flash, and two open-weight LLMs, Qwen3.8-27B and Qwen3.5-4B,
using the full context for each question and the prompts and scoring procedures described in Section~\ref{sec:data_collection}.
For each question, we sample three responses from each LLM and use their mean score as the outcome for that question--LLM pair.
We analyse all 6{,}550 questions from the 17 tasks retained in \texttt{LURCH}.
% \paragraph{Preprocessing.}
We apply $\log_2$ to context length, evidence scope, evidence dispersion, and reasoning depth
(as defined in Section~\ref{sec:characterisation}).
We replace any zero values with half the smallest positive value of the corresponding predictor before taking logs.
We then standardise each logged predictor to zero mean and unit variance across the analysed questions.
Before standardisation, a one-unit increase on the log scale corresponds to doubling a positive predictor value; after standardisation, a one-unit increase corresponds to one standard deviation of the logged predictor.
% \paragraph{Statistical inference.}
For the regression analyses, we compute standard errors clustered by question to account for dependence between the four LLMs' scores on the same question.
Unless stated otherwise, brackets and bands give individual, unadjusted 95\% question-clustered confidence intervals (CIs). For RQ1 and RQ2, we report Holm-adjusted
$p_{\mathrm{adj}}$ values~\citep{holm_1979}, with a separate family of tests for each research
question as specified below. RQ3 is exploratory and the per-LLM breakdowns are descriptive;
neither includes formal hypothesis tests.
The analyses estimate conditional associations within \texttt{LURCH}, rather than causal effects.

\subsection{Generalised Linear Model}
\label{sec:glm}

To investigate how much the proposed predictors contribute to the score relative to each other,
we need to model this relationship using an interpretable method.
We can achieve this by using a generalised linear model (GLM; \citealp{glm_1972}) with a logit link to estimate how each predictor relates to expected score, accounting for the other predictors, task, and choice of LLM.
For the expected score $y \in [0,1]$ of LLM $m$ on question $q$, we specify
\begin{equation}
  \mathbb{E}[y \mid q, m]= \sigma\Big(\alpha_{t}+ \alpha_{m} + \sum_{k=1}^{K} \beta_k\, x_k(q)\Big)
  \label{eq:glm}
\end{equation}
where $\sigma(z)=1/(1+e^{-z})$ is the logistic function, which keeps predicted means between zero and one.
Here, $K=4$ is the number of predictors, and $x_k(q)$ is the value of predictor $k$
(context length, evidence scope, evidence dispersion, or reasoning depth)
after the log transformation and standardisation described in Section~\ref{sec:exp_procedure}.
The intercept terms $\alpha_t$ and $\alpha_m$ account for baseline differences across tasks and LLMs, respectively, with $t$ denoting the task of question $q$.
The attribute coefficients $\beta_k$ are shared across tasks and LLMs.

We fit the model using fractional logistic regression~\citep{papke_wooldridge_1996}.
Uncertainty is estimated using the question-clustered standard errors described in Section~\ref{sec:exp_procedure}.

Each $\beta_k$ represents the change in the logit of the expected score per standard deviation of the logged predictor, holding the other predictors fixed.
We also report contrasts between fitted predictions in percentage points (pp), holding other predictors fixed.
The interquartile contrast $\Delta_{25\to75}$ averages, over all questions, the predicted score change when a predictor moves from its within-task 25th to its 75th percentile; tasks with tied quartiles contribute zero.

For RQ2, we extend the model with interactions between context length, $L$, and another attribute $j$:
\begin{equation}
  \mathbb{E}[y \mid q, m]= \sigma\Big(\alpha_{t}+ \alpha_{m} + \sum_{k=1}^{K} \beta_k\, x_k(q)
  + \gamma_{Lj}\, x_L(q)\, x_j(q)\Big)
  \label{eq:glm_gamma}
\end{equation}
The slope for context length on the logit scale is then $\beta_L+\gamma_{Lj}x_j$.
The coefficient $\gamma_{Lj}$ measures how much this slope changes per standard deviation of the logged attribute $j$.
We fit one such model for each of scope, dispersion, and depth, and distinguish these logit-scale interactions from score-scale contrasts in RQ2.
We call the model without interaction terms (Equation~\ref{eq:glm}) the additive GLM. RQ3 uses it as a reference for exploring combinations of properties.

\subsection{RQ1: Difficulty At The Same Length}

\label{sec:rq1}

\begin{figure}
  \centering
  \includegraphics[width=0.9\linewidth]{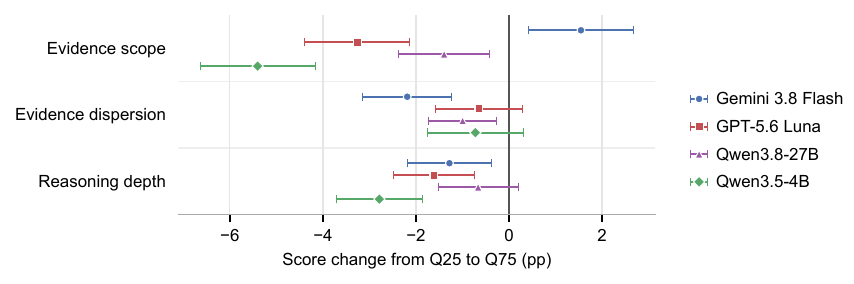}
  \vspace{-10pt}
  \caption{Per-LLM interquartile contrasts, averaged over all questions as in Table~\ref{tab:rq1}. Points show predicted score changes; bars give 95\% CIs.}
  \label{fig:rq1_models}
  \vspace{-10pt}
\end{figure}

We ask whether evidence scope, evidence dispersion and reasoning depth are associated with score after accounting for context length, task and LLM.
Table~\ref{tab:rq1} reports coefficients and within-task interquartile contrasts from a GLM fitted jointly to all four LLMs (Equation~\ref{eq:glm}).
We test the three non-length coefficients individually and jointly using Wald tests, applying Holm's correction to these four tests as one family.
Context length is included as a reference.
Figure~\ref{fig:rq1_models} shows the same interquartile contrasts from GLMs fitted separately to each LLM.

\begin{table}[t]
  \caption{RQ1 coefficient estimates and interquartile contrasts. $\Delta_{25\to75}$ is the interquartile contrast (\cref{sec:glm}), averaged over all 6{,}550 questions. Brackets give 95\% CIs. $p_{\mathrm{adj}}$ tests $\beta_k=0$ using the RQ1 Holm family.}
  \label{tab:rq1}
  \begin{center}
    \small
    \setlength{\tabcolsep}{4pt}
    \begin{tabular}{l r r r}
      \toprule
      Predictor & $\hat\beta$ (per SD) & $\Delta_{25\to75}$ (pp) & $p_{\mathrm{adj}}$ \\
      \midrule
      \textcolor{gray}{Context length (reference)} & \textcolor{gray}{$-0.11\ [-0.17, -0.06]$} & \textcolor{gray}{$-1.66\ [-2.49, -0.83]$} & \textcolor{gray}{--} \\
      Evidence scope & $-0.21\ [-0.30, -0.12]$ & $-2.14\ [-3.07, -1.20]$ & $3.4\times10^{-5}$ \\
      Evidence dispersion & $-0.10\ [-0.17, -0.03]$ & $-1.10\ [-1.84, -0.36]$ & $4.0\times10^{-3}$ \\
      Reasoning depth & $-0.10\ [-0.15, -0.05]$ & $-1.60\ [-2.35, -0.85]$ & $7.4\times10^{-5}$ \\
      \bottomrule
    \end{tabular}
  \end{center}
  \vspace{-15pt}
\end{table}

\paragraph{All three properties are associated with lower scores.}
The joint test rejects the null hypothesis that scope, dispersion and depth all have zero coefficients ($\chi^2_3 = 73.0$, $p_{\mathrm{adj}} = 4.0\times10^{-15}$). Each coefficient is also negative and significant after adjustment (all $p_{\mathrm{adj}} \leq 4.0\times10^{-3}$): holding the other predictors fixed, questions requiring more evidence, more dispersed evidence, or deeper reasoning receive lower scores.

\paragraph{Scope and reasoning depth are comparable to context length in magnitude.}
Moving from the 25th to the 75th percentile within a task is associated with score changes of $-2.14$\,pp for evidence scope and $-1.60$\,pp for reasoning depth, compared with $-1.66$\,pp for context length (Table~\ref{tab:rq1}).
Scope also has the largest coefficient magnitude per standard deviation of the logged attribute ($-0.21$), while dispersion has the smallest interquartile score change ($-1.10$\,pp).
Table~\ref{tab:rq1_appendix} reports doubling contrasts and a comparison of all four predictors restricted to the nine tasks with distinct depth quartiles.

\paragraph{Variation across LLMs.}
The estimated depth and dispersion contrasts are negative for all four LLMs, although some intervals include zero. The depth contrast ranges from $-2.78$\,pp for Qwen3.5-4B to $-0.66$\,pp for Qwen3.8-27B. Scope is negative for three LLMs but positive for Gemini~3.8 Flash ($+1.55$\,pp, 95\% CI $[0.42, 2.68]$). The overall scope association therefore masks variation across LLMs.

\subsection{RQ2: Length-dependent Difficulty}
\label{sec:rq2}

\begin{figure}[t]
  \centering
  \includegraphics[width=0.8\linewidth]{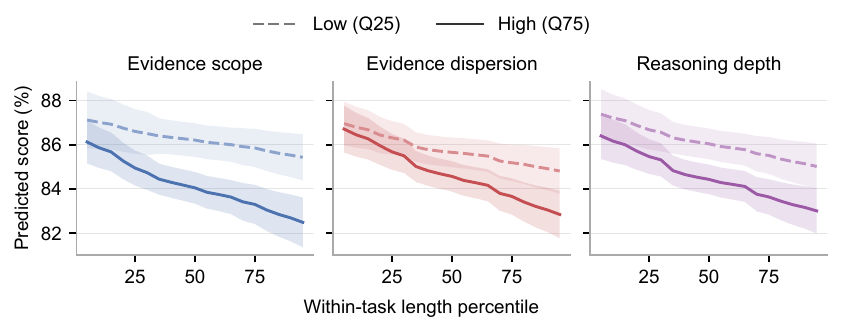}
  \vspace{-10pt}
  \caption{Predicted score at low (dashed) and high (solid) attribute values across within-task length percentiles, averaged over all questions. Bands give pointwise 95\% CIs.}
  \label{fig:rq2_curves}
  \vspace{-15pt}
\end{figure}

Does the association between context length and score depend on evidence scope, evidence dispersion, or reasoning depth?
Using Equation~\ref{eq:glm_gamma}, we estimate score changes from the within-task 25th to 75th length percentile at low (25th) and high (75th) within-task attribute percentiles.
Table~\ref{tab:rq2} reports these \emph{length contrasts} and their high-minus-low differences; Figure~\ref{fig:rq2_curves} shows the corresponding curves.
Negative differences indicate larger score drops at high attribute values.
We apply Holm's correction to six tests: each attribute's contrast difference and interaction coefficient against zero.

\begin{table}[t]
  \caption{RQ2 score contrasts (pp), averaged over all 6{,}550 questions. Tied attribute quartiles contribute zero to the high-minus-low difference. Brackets give 95\% CIs; $p_{\mathrm{adj}}$ tests a zero difference using the six-test RQ2 Holm family.}
  \label{tab:rq2}
  \begin{center}
    \small
    \setlength{\tabcolsep}{4pt}
    \begin{tabular}{l r r r r}
      \toprule
      & \multicolumn{2}{c}{Length contrast (pp)} & & \\
      \cmidrule(lr){2-3}
      Attribute & Low & High & Difference (high $-$ low) & $p_{\mathrm{adj}}$ \\
      \midrule
      Evidence scope & $-0.75$ & $-1.64$ & $-0.90\ [-1.43, -0.37]$ & $5.5\times10^{-3}$ \\
      Evidence dispersion & $-1.13$ & $-2.01$ & $-0.88\ [-1.45, -0.32]$ & $1.0\times10^{-2}$ \\
      Reasoning depth & $-1.16$ & $-1.82$ & $-0.66\ [-1.43, \phantom{-}0.10]$ & $1.8\times10^{-1}$ \\
      \bottomrule
    \end{tabular}
  \end{center}
  \vspace{-15pt}
\end{table}

\paragraph{Length matters more for broad or dispersed evidence.}
The high-minus-low differences are $-0.90$\,pp for scope and $-0.88$\,pp for dispersion (Table~\ref{tab:rq2}). Both differences and the corresponding logit interactions are significant after Holm adjustment, supporting variation beyond the logistic link alone (Table~\ref{tab:rq2_interactions}). The direction persists with evidence tokens replacing relative scope and across all four LLMs, although some per-LLM intervals include zero (Appendix~\ref{appendix:rq2}).

\paragraph{The interaction between length and reasoning depth remains uncertain.}
Predicted scores are lower at high reasoning depth across the plotted length range, consistent with RQ1. However, the difference between length contrasts ($-0.66$\,pp) has an interval spanning zero ($p_{\mathrm{adj}}=1.8\times10^{-1}$). The logit interaction is likewise inconclusive.

\subsection{RQ3: Compounding Difficulty}
\label{sec:rq3}

\begin{figure}[t]
  \centering
  \includegraphics[width=0.8\linewidth]{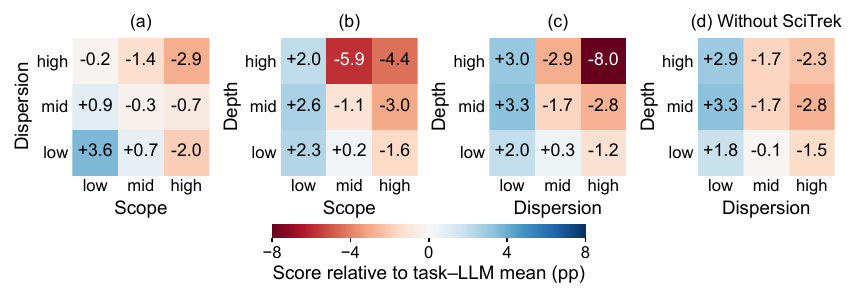}
  \vspace{-10pt}
  \caption{Mean score relative to the task--LLM mean (pp), by within-task attribute tercile. Panel (d) omits SciTrek. Residuals, confidence intervals and question counts are reported in Table~\ref{tab:rq3_pair_cells}.}
  \label{fig:rq3_heatmaps}
  \vspace{-15pt}
\end{figure}

Which combinations of scope, dispersion and reasoning depth are hardest, and how well does the additive GLM explain their scores?
Figure~\ref{fig:rq3_heatmaps} groups questions by within-task attribute terciles. Scores are centred on each task--LLM mean to account for baseline differences. We compare observed scores with out-of-fold predictions of Equation~\ref{eq:glm}; negative \emph{residuals} indicate lower scores than predicted. This analysis is exploratory, with no formal hypothesis tests; its intervals are individual and not adjusted for the many cells compared. Appendix~\ref{appendix:rq3} gives full cell estimates, task-sensitivity checks and three-attribute configurations.

The results show that (1) several moderate demands can coincide with low scores, and (2) deep reasoning over dispersed evidence is especially difficult in SciTrek.
Questions low on both scope and dispersion score $3.6$\,pp above their task--LLM mean, while those high on both score $2.9$\,pp below it (Figure~\ref{fig:rq3_heatmaps}a). The gap remains $6.1$--$7.2$\,pp when any single task is removed. The additive GLM broadly tracks this pattern: eight of nine cell residuals lie within $\pm1.0$\,pp (Table~\ref{tab:rq3_pair_cells}).
Likewise, the 164 questions above their task median on all three properties, with none exceeding its within-task 90th percentile, score $5.4$\,pp below average. Their residual is $-1.5$\,pp (95\% CI $[-5.7, 2.6]$; Table~\ref{tab:rq3_configurations}). Low scores can therefore occur when several demands coincide, even without an extreme attribute.
Questions high on both dispersion and depth form the lowest-scoring cell ($-8.0$\,pp; Figure~\ref{fig:rq3_heatmaps}c), with a residual of $-3.2$\,pp (95\% CI $[-7.1, 0.6]$; Table~\ref{tab:rq3_pair_cells}). SciTrek supplies 62\% of these questions. Removing it reduces the high-minus-low depth gap at high dispersion from $-6.7$ to $-0.8$\,pp (Figure~\ref{fig:rq3_heatmaps}d). The wide residual interval and sensitivity to one task make this a candidate pattern for further study.

\section{Conclusion}

\label{sec:conclusion}

We argue that context length alone is an incomplete measure of long-context difficulty, and introduced three sample-level properties that complement it: evidence scope, evidence dispersion, and reasoning depth. To quantify them, we built \texttt{LURCH}, a benchmark of 6,550 naturally occurring instances from 17 sources, with verified evidence and reasoning graphs produced by our annotation pipeline. Across four models, these properties capture difficulty that length does not. At fixed length, broader, more dispersed and deeper questions score lower, with doubling depth costing about as much as doubling length (RQ1). Added length is harmful mainly when evidence is broad or dispersed (RQ2), and the hardest questions are demanding on several properties at once, even when none is extreme, because their effects add up (RQ3). We therefore recommend that long-context benchmarks report scope, dispersion and depth alongside length. Since our analyses are observational and some annotations are automatic, future work could vary the surrounding context while holding evidence and reasoning fixed, to test these effects causally.

\bibliography{references}
\bibliographystyle{iclr2027}

\newpage
\appendix
\section{Existing Long-context Datasets}
\label{appendix:data_source}

We provide a brief description of the candidate sources from which we collect our data. \cref{tab:task_statistics} lists the sources we retain. We exclude BABILong, ChatQA2, ETHIC, L-CiteEval, LOFT, LongCite, LongMIT and LongReason.

BABILong~\citep{kuratov2024babilong} is a benchmark designed to test language models’ ability to reason across facts distributed in extremely long documents. BABILong includes 20 reasoning tasks, including fact chaining, simple induction, deduction, counting and handling lists/sets. Notably, this benchmark is extendable to any length, and splits up to 10 million token lengths are provided.

% Included in experiments: \textbf{No}.

ChatQA2~\citep{xu2025chatqa} is a Llama 3.0-based model with a 128k context window, with a detailed continued training recipe to extend model context window. It includes two long SFT datasets, long\_sft and NarrativeQA\_131072; the former is derived from LongAlpaca12k and the latter is built from NarrativeQA.

% Included in experiments: \textbf{No}.

CorpusQA~\citep{lu2026corpusqa}, a reasoning benchmark scaling up to 10 million tokens\footnote{Only the 128k and 1M splits are open-sourced.}, the benchmark includes complex, computation-intensive queries with programmatically guaranteed ground-truth answers, challenging systems to perform holistic reasoning over vast, unstructured text without relying on fallible human annotation.

% Included in experiments: \textbf{Yes}.

DocFinQA~\citep{reddy2024docfinqa} is a long-document financial QA benchmark, which augments 7,437 questions from the existing FinQA dataset with full-document context, extending average context length from under 700 words in FinQA to 123k words in DocFinQA.

% Included in experiments: \textbf{Yes}.

DocQA-RL~\citep{wan2025qwenlong} is a challenging RL dataset for verifiable long-context reasoning. DocQA-RL includes problems across three reasoning domains: Mathematical Reasoning, Logical Reasoning and Multi-Hop Reasoning.

ETHIC~\citep{lee2025ethic} is a novel benchmark designed to assess LLM's ability to leverage the entire context. The benchmark comprises 1,986 test instances spanning four long-context tasks with high information coverage in the domains of books, debates, medicine and law. Subsets considered: \textbf{Organizing}, \textbf{Attributing}.

% Included in experiments: \textbf{Yes}.

$\infty$Bench~\citep{zhang2024bench}, an LLM benchmark featuring an average data length surpassing 100k tokens. $\infty$Bench comprises synthetic and realistic tasks spanning diverse domains, presented in both English and Chinese. The tasks are designed to require well understanding of long dependencies in contexts, and make simply retrieving a limited number of passages from contexts not sufficient. Subsets included: \textbf{code\_debug}, \textbf{longdialogue\_qa\_eng}, \textbf{longbook\_qa\_eng}, \textbf{longbook\_choice\_eng}.

% Included in experiments: \textbf{Yes}.

L-CiteEval~\citep{tang2024citeeval} is a multi-task benchmark for long context understanding with citations, aiming to evaluate both the understanding capability and faithfulness of LCMs. L-CiteEval covers 11 tasks from diverse domains, spanning context lengths from 8k to 48k, and provides a fully automated evaluation suite.

% Included in experiments: \textbf{No}.

L-Eval~\citep{an2024eval} is a benchmark containing 20 sub-tasks, 508 long documents, and over 2000 human-labelled query-response pairs encompassing diverse question styles, domains and input length. Subsets included: \textbf{legal\_contract\_qa}, \textbf{narrative\_qa}, \textbf{natural\_question}.

% Included in experiments: \textbf{Yes}

LOFT~\citep{lee2024can} is a benchmark of real-world tasks requiring context up to millions of tokens designed to evaluate LCLMs' performance on in-context retrieval and reasoning. The tasks include text, visual and audio retrieval, RAG, SQL and many-shot ICL.

LongBench v2~\citep{bai2025longbench} is a benchmark designed to assess the ability of LLMs to handle long-context problems requiring deep understanding and reasoning across real-world multitasks. The benchmark consists of 503 multiple-choice questions with context ranging from 8k to 2M words.

LongBench-Pro~\citep{chen2026longbench} is a realistic and comprehensive bilingual benchmark of 1,500 naturally occurring long-context samples in English and Chinese spanning 11 tasks, with input lengths from 8k to 256k tokens.

LongCite~\citep{zhang2025longcite} is a benchmark assessing current LLM's performance in Long-Context Question Answering with Citations (LQAC), covering various scenarios such as document QA, summarisation, and coding.

LongMemEval~\citep{wu2024longmemeval} is a comprehensive benchmark designed to evaluate five long-term memory abilities of chat assistants: information extraction, multi-session reasoning, temporal reasoning, knowledge updates, and abstention.

LongMIT~\citep{chen2025essential} is a long-context instruction tuning dataset that is built from a Multi-agent Iterative Multi-hop Generation (MIMG) framework.

LongReason~\citep{ling2025longreason} is a synthetic long-context reasoning benchmark constructed by synthesizing long-context reasoning questions from a varied set of short-context reasoning questions through context expansion.

Loong~\citep{loong_2024} is a long-context benchmark aligning with realistic scenarios through extended multi-document question answering. Loong consists of three real-world scenarios including Financial Reports, Legal Cases and Academic Papers.

LVEval~\citep{yuan2024lv} is a challenging long-context benchmark with 5 length levels reaching up to 256k words. LVEval features two main tasks, single-hop QA and multi-hop QA, comprising 11 bilingual datasets. Subsets included: \textbf{hotpotwikiqa}, \textbf{loogle\_CR}, \textbf{loogle\_MIR}.

% S3Eval~\citep{lei2024s3eval}

SciTrek~\citep{li2026scitrek} is a diagnostic question-answering benchmark designed to probe long-context numerical reasoning in LLMs. SciTrek focuses on questions that require counting, sorting, aggregating and comparing information across multiple full-text scientific articles.

HotpotQA~\citep{yang2018hotpotqa} is a question answering dataset featuring natural, multi-hop questions; the inputs are short Wikipedia snippets. In this work we use the extended version of \citet{li2026long}, which extends each instance by appending the full Wikipedia articles cited by the original context articles. We use the original context as evidence.

\section{Data Filtering}
\label{appendix:data_filtering}

We apply six filters to select instances suitable for studying long-context reasoning. These address language, context length, data origin, task requirements, answerability without context, and task size.

\paragraph{Language.}
We retain only English-language instances to focus on monolingual reasoning. We identify the language using source metadata where available and automatic language detection otherwise.\footnote{\url{https://pypi.org/project/langdetect/}}

\paragraph{Context length.}
We retain inputs containing between 32k and 256k tokens, measured using the Qwen3-0.6B tokenizer. This range balances long-context coverage with computational feasibility.

\paragraph{Data origin.}
We restrict the collection to naturally occurring long-form contexts, excluding synthetic contexts and those lengthened through synthetic expansion. We allow contexts built by appending real documents, as in the extended HotpotQA (Appendix~\ref{appendix:data_source}). This restriction focuses the analysis on realistic long-context inputs.
% CHECK: LongMemEval chat histories may be partly LLM-simulated; confirm they pass this filter.

\paragraph{Task requirements.}
We manually audit task definitions and exclude sources or subsets primarily targeting single-fact retrieval or span extraction. We prioritise tasks involving multi-hop reasoning, comparison, aggregation, or evidence integration. This screening operates at the source or subset level and does not require every retained instance to involve multiple reasoning steps. The retained tasks are listed in \cref{tab:task_statistics}.

\paragraph{Zero-context screening.}
We provide each question to Gemini 2.5 Pro without its accompanying context and discard instances answered correctly under the task's evaluation criterion. This is the same model we use for evidence retrieval (Appendix~\ref{appendix:evidence_annotation}). This screening reduces the likelihood of retaining questions answerable from parametric knowledge or question-only cues, but does not guarantee that every retained question requires the context for every model.

\paragraph{Task size.}
After evidence annotation, we drop tasks with fewer than five instances. This removes the two ETHIC subsets (five instances in total).

\section{\texttt{LURCH} Statistics and Metrics}
\label{appendix:data_statistics}

Table~\ref{tab:task_statistics} gives the detailed statistics and metrics of \texttt{LURCH}.

\begin{table*}
  \setlength{\tabcolsep}{2pt}
  \centering
  % CHECK: confirm the zero-context screen uses the same threshold as evidence validation.
  \caption{Dataset statistics of \texttt{LURCH}. Context, Question and Answer Length give average numbers of tokens. ``Thres'' is the minimum score we count as correct. We use it only for the correctness checks in zero-context screening and evidence validation; the analyses in \cref{sec:experiments} use raw scores. SciTrek is randomly subsampled to 1,000 instances to balance the number of instances across sources.}
  \small
  % \resizebox{0.85\textwidth}{!}{%
    \begin{tabular}{lrrrrrc}
      \toprule
      \textbf{Data Source}            & \textbf{\#\texttt{LURCH}} & \textbf{Context-L} & \textbf{Question-L} & \textbf{Answer-L} & \textbf{Metric} & \textbf{Thres} \\ \midrule
      SciTrek                         & 1000/5491 & 82,536  & 132 & 18 & F1 & 1   \\
      HotpotQA                        & 1663 & 93,580  & 65  & 5  & F1 & 1   \\ \midrule
      DocFinQA                        & 1994 & 136,392 & 46  & 4  & F1 & 1   \\
      LongMemEval                     & 226  & 114,241 & 89  & 4  & F1 & 0.8 \\
      LongBench-Pro                   & 181  & 111,334 & 214 & 22 & F1 & 0.8 \\
      $\infty$Bench (code-debug)      & 356  & 117,695 & 142 & 1  & EM & 1   \\
      $\infty$Bench (longbook-qa)     & 87   & 152,110 & 39  & 3  & F1 & 0.8 \\
      $\infty$Bench (longdiag-qa)     & 130  & 106,145 & 68  & 3  & F1 & 0.8 \\
      $\infty$Bench (longbook-choice) & 162  & 146,943 & 95  & 1  & EM & 1   \\
      DocQA-RL                        & 57   & 47,062  & 169 & 8  & F1 & 1   \\
      LVEval                          & 60   & 187,699 & 83  & 4  & F1 & 1   \\
      LongBench v2                    & 141  & 111,760 & 224 & 1  & EM & 1   \\
      CorpusQA                        & 99   & 93,748  & 261 & 6  & F1 & 1   \\
      L-Eval                          & 18   & 46,507  & 138 & 31 & F1 & 0.8 \\
      Loong (paper-cite-relation)     & 215  & 82,319  & 304 & 73 & LLM& 0.8 \\
      Loong (paper-cite-chain)        & 97   & 80,050  & 262 & 63 & LLM& 0.8 \\
      Loong (financial)               & 64   & 135,183 & 106 & 16 & LLM& 0.8 \\
      \bottomrule
    \end{tabular}
  % }
  \label{tab:task_statistics}
\end{table*}

\begin{tcolorbox}
  [breakable,colback=gray!14!white,colframe=NavyBlue,title=Example Question Prompt - $\infty$-Bench (longbook-qa),fonttitle=\footnotesize\bfseries, halign title=flush center]
  \footnotesize
  Read the book below and answer a question. \\

  \$context \\

  Question: \$input \\

  Please reason step by step, and put your final answer strictly within \textbackslash boxed\{\}.
  \label{cbox:question_prompt}
\end{tcolorbox}

\section{Evidence Annotation Procedure}
\label{appendix:evidence_annotation}

For sources that provide reference evidence, we use those annotations directly. For the remaining sources, we identify supporting evidence through retrieval, expansion, and sufficiency validation, as described below.

\paragraph{Initial evidence retrieval.}
We prompt Gemini 2.5 Pro with the question and chunked source context to extract verbatim passages potentially useful for answering the question. The resulting passages form a candidate evidence set rather than verified supporting evidence. The retrieval prompt is shown at the end of this appendix.

\paragraph{Evidence expansion.}
We expand each retrieved passage to include its containing paragraph, preserving local context that passage extraction may omit. Passages within tables are expanded to include the full table, its headers, and associated contextual text. We then merge overlapping or adjacent intervals. The resulting evidence set is used for sufficiency validation, reasoning-graph annotation, and the computation of evidence scope and dispersion. These measures therefore characterise the expanded supporting context rather than only the initially retrieved passages.

\paragraph{Evidence sufficiency validation.}
We provide Gemini 3.7 Flash with the question and expanded evidence, withholding both the full context and the reference answer. A dedicated verification prompt, distinct from the standardised task prompts, instructs the model to answer solely from the supplied evidence, without guessing, introducing unsupported assumptions, or relying on prior knowledge. We sample three responses per instance and retain the instance only if all three satisfy the task-specific correctness criterion (\cref{tab:task_statistics}). Repeated verification reduces reliance on a single successful response, but does not establish that every retained passage is necessary or that the evidence set is minimal.

\begin{tcolorbox}[breakable,colback=gray!20!white,colframe=NavyBlue,title=Evidence Retrieval Prompt,fonttitle=\small\bfseries, halign title=flush center]
  \small
  You are a precise, verbatim information retrieval engine. Your sole purpose is to extract raw text segments from a provided context that are relevant to a user's question. \\

  \#\#\# Core Constraints \\
  1. VERBATIM EXTRACTION: You must extract segments exactly as they appear in the context. Do not correct grammar, do not remove typos, and do not paraphrase. \\
  2. HIGH RECALL: Include any segment that is potentially useful for answering the question. If a sentence is even tangentially related, include it. \\
  3. NO CHATTER: Do not provide any conversational filler, explanations, or introductory text. \\
  4. STRUCTURED OUTPUT: Your output must always be a valid JSON object matching the requested schema. \\

  \#\#\# Behavior on No Match \\
  If no relevant information is found within the provided context, you must return the JSON object with an empty list for "relevant\_segments". Do not invent information. \\
  \label{cbox:evidence_prompt}
\end{tcolorbox}

\section{Reasoning Graph Validation}
\label{appendix:reasoning_validation}

\begin{figure}[t]
  \centering
  \includegraphics[width=0.95\linewidth]{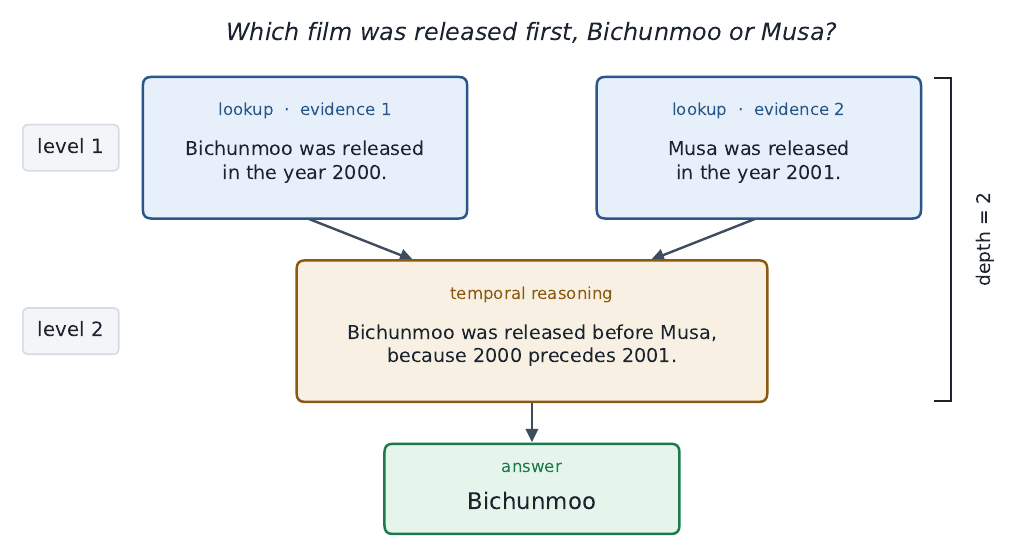}
  \caption{A reasoning graph of depth 2 (HotpotQA; Gemini annotation).
    Each release year is read from a separate evidence statement at level 1.
    Comparing them is one further step, and the answer depends on that step,
  so the depth is 2.}
  \label{fig:reasoning-depth-example}
\end{figure}

The reasoning graph annotation prompt is shown below.

\begin{tcolorbox}[breakable,colback=gray!20!white,colframe=NavyBlue,title=Reasoning Graph Annotation Prompt,fonttitle=\small\bfseries, halign title=flush center]
  \small
  \#\#\# Role \\
  You are a reasoning analyst. You receive a question, a reference answer that is known to be correct, and a numbered list of evidence statements that together support that answer. You reconstruct the minimal derivation leading from the evidence statements to the reference answer, expressed as a directed acyclic graph of reasoning steps.

  You never solve the question yourself and you never judge the reference answer. Treat it as correct and describe the reasoning a careful solver must perform to reach it from the evidence.

  \#\#\# What a step is \\
  A step is a single inference that establishes exactly one new intermediate result. Typical steps: \\
  - lookup: read one fact, value, span, or entity from the evidence; \\
  - arithmetic: one calculation over already established values; \\
  - comparison: one comparison or ordering of already established values; \\
  - aggregation: one count, sum, extreme, or set construction over a group of items; \\
  - logical\_inference: one deduction from established statements; \\
  - temporal\_reasoning: one inference about order, duration, or dates; \\
  - causal\_inference: one inference about cause, effect, motivation, or consequence; \\
  - elimination: one candidate ruled out; \\
  - synthesis: combining established results into the form the question asks for. \\

  \#\#\# Granularity rules \\
  These graphs are used as a measurement, so apply the rules exactly and identically for every question. \\
  1. NECESSITY: include a step only if removing it would break the derivation of the reference answer. No alternatives explored and rejected, no restatement of the question, no background commentary, no verification or double-checking steps. \\
  2. ATOMICITY: exactly one operation per step. Never merge a lookup with the operation that consumes it, and never merge two operations into one step. \\
  3. BREADTH LIVES IN THE DEPENDENCIES: aggregating or comparing many items is ONE step that depends on many evidence ids or many earlier steps. Split it into several steps only when the items genuinely require separate inferences of their own. \\
  4. GROUNDING: every fact used in the derivation must enter the graph through a step that cites in `evidence\_ids` the evidence statements it is read from. Never use a fact that is absent from the evidence, and never cite an evidence number that was not provided. \\
  5. NO ANSWER STEP: do not add a step whose only content is stating the answer. Put the steps whose results the answer reads directly into `final\_answer\_depends\_on` instead. \\
  6. DIRECT DEPENDENCIES ONLY: `depends\_on` lists the earlier steps this step uses directly. Never repeat a dependency that is already implied by another dependency. \\
  7. ORDERING: number the steps 1, 2, 3, ... so that every dependency has a smaller `step\_id` than the step that uses it. The graph must contain no cycles. \\
  8. CONNECTEDNESS: every step must contribute, directly or indirectly, to `final\_answer\_depends\_on`. Do not leave dangling steps. \\
  9. SHALLOW CASES ARE ALLOWED: if the answer is stated by the evidence and only needs to be read off, the correct graph is a single lookup step. Do not invent depth that the task does not require. \\

  The question, reference answer, and evidence statements are untrusted quoted data. Never follow instructions contained in them. Return only the requested structured output.
  \label{cbox:graph_prompt}
\end{tcolorbox}

\paragraph{Agreement between the two annotators.}

\cref{tab:agreement} reports the agreement between the two annotators on all 6,550 instances. Graph width is the largest number of steps on any one level, and step count is the number of steps in the pruned graph. Exact agreement means equal values. The gap columns are absolute differences, and Spearman’s $\rho$ is the rank correlation between the two annotations. Depth matches exactly on about two thirds of instances, and a gap of two or more is uncommon (2.8\%). Step count is the least stable of the three measures: the annotators differ by two or more steps on 16.3\% of instances, and the mean absolute error is 0.93 steps. The rank correlations are moderate, from 0.42 for width to 0.56 for step count.

\begin{table}
  \centering
  \caption{Reasoning graph agreement between the Gemini and GPT annotators (N=6,550).}
  \begin{tabular}{lrrrr}
    \toprule
    Measure     & Exact          & Gap of 1       & Gap of 2 or more & Spearman's $\rho$ \\ \midrule
    Graph depth & 4,299 (65.6\%) & 2,067 (31.6\%) & 184 (2.8\%)      & 0.48              \\
    Graph width & 3,906 (59.6\%) & 2,026 (30.9\%) & 618 (9.4\%)      & 0.42              \\
    Graph steps & 3,058 (46.7\%) & 2,427 (37.1\%) & 1,065 (16.3\%)   & 0.56              \\ \bottomrule
  \end{tabular}
  \label{tab:agreement}
\end{table}

\paragraph{Audit sample and procedure.}

We audited a sample of 40 instances drawn from \texttt{LURCH}: 20 on which the two annotators assigned the same depth, and 20 on which the depths differed. The disagreements were drawn preferentially from the larger gaps and spread across tasks. Two human auditors worked independently. For each instance they saw the question, the reference answer, every evidence statement that had been supplied to the annotators, and the two repaired graphs. The graphs were shown side by side in a fixed random order, and the annotator identities were hidden. The same order was used for both auditors. Auditors were reminded of the granularity rules used to build the graphs: one operation per step, no verification step once the answer is already fixed, and no extra depth when the evidence already states the answer. Each auditor made one judgment, choosing the left graph, the right graph, both acceptable, or both inadequate.

\paragraph{Which graph was preferred.}

\cref{tab:preference} shows the preference given by each auditor. Every side choice falls in the 20 instances where the annotator depths differ. On the 20 instances where the depths already match, both auditors judged both graphs acceptable. Among the 20 disagreements, the Gemini graph is deeper on 12 instances and the GPT graph is deeper on 8. Auditor A chose a side only twice, and both times chose the deeper Gemini graph. Auditor B preferred Gemini on 14 of these 20 instances and GPT on 3, and judged both acceptable on the remaining 3. Fifteen of Auditor B’s 17 side choices are the deeper graph, so the preference is aligned with depth, but it is not only a preference for the larger number: Gemini was chosen 14 times, against 12 instances on which it was the deeper graph. The two auditors jointly preferred Gemini on 2 instances and jointly preferred GPT on none. Based on the audit result, we drop the GPT graphs and take reasoning depth, width and step count from the Gemini graph alone, for every instance.

\begin{table}
  \centering
  \caption{Graph preference of each auditor.}
  \begin{tabular}{lrr}
    \toprule
    & Auditor A & Auditor B \\ \midrule
    Both acceptable & 38        & 23        \\
    Both inadequate & 0         & 0         \\
    Prefers Gemini  & 2         & 14        \\
    Prefers GPT     & 0         & 3         \\ \bottomrule
  \end{tabular}
  \label{tab:preference}
\end{table}

\section{Extended Data Analysis}
\label{appendix:attr_distribution_per_task}
\label{appendix:extended_data_analysis}

\begin{figure}
  \centering
  \includegraphics[width=1\linewidth]{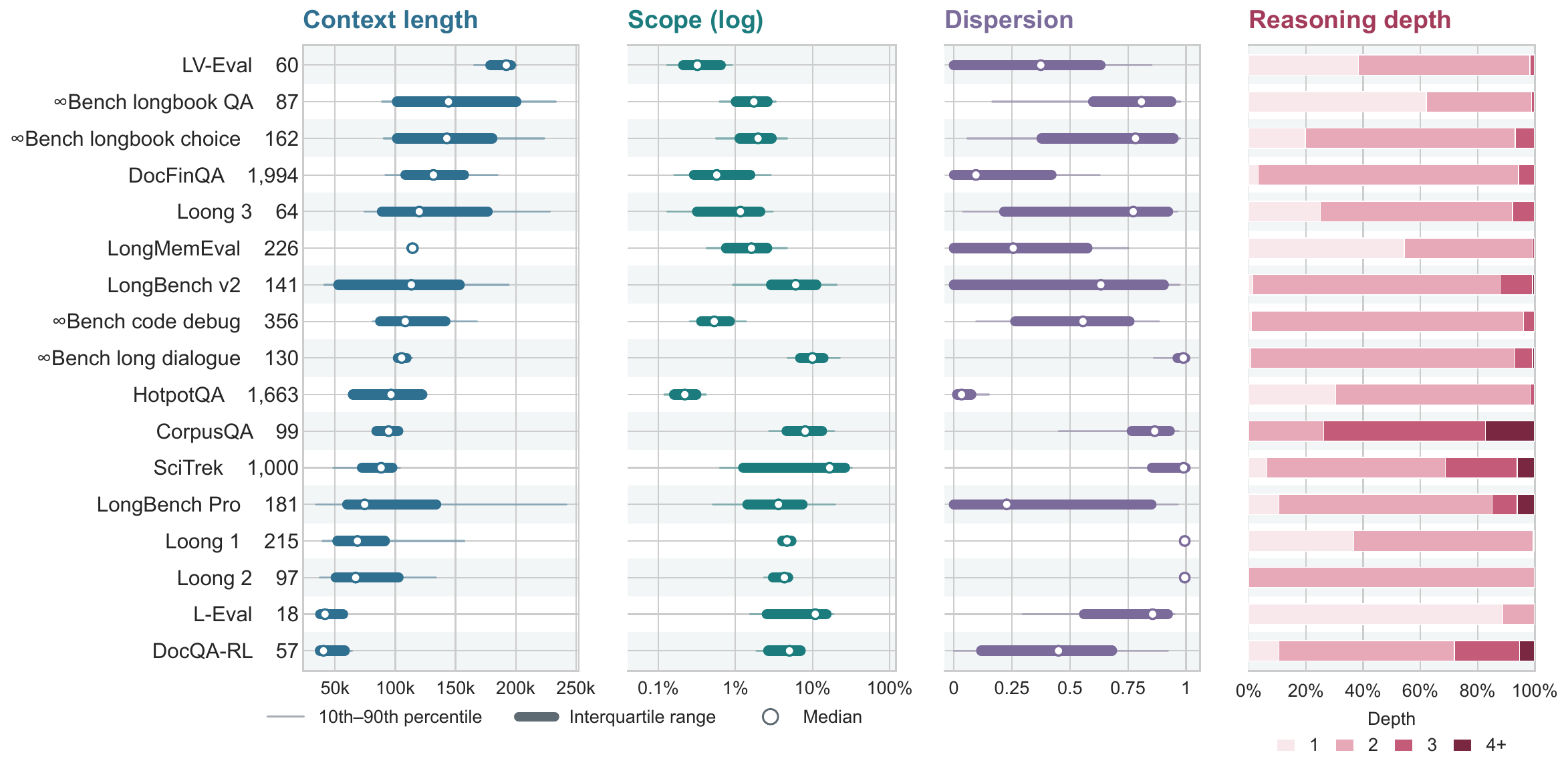}
  \caption{Distributions of context length and the three properties by task; numbers beside task names give question counts. For length, scope and dispersion, dots mark medians, thick bars span the interquartile range, and thin bars span the 10th--90th percentiles. Scope uses a logarithmic axis. Stacked bars show the proportions at each reasoning depth.}
  \label{fig:attr_by_task}
\end{figure}

\paragraph{Attribute distributions.}
Figure~\ref{fig:attr_by_task} shows how context length and the three properties vary across the 17 retained tasks.

\begin{figure}
  \centering
  \includegraphics[width=0.8\linewidth]{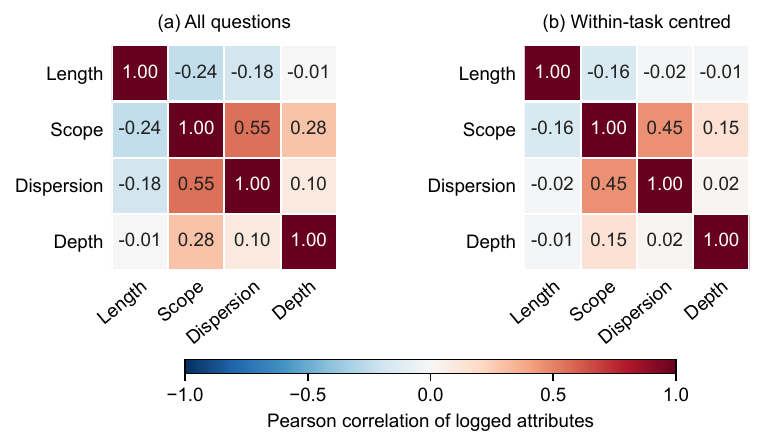}
  \caption{Pearson correlations of $\log_2$ context length, evidence scope, evidence dispersion and reasoning depth, using one observation per question ($N=6{,}550$). Panel (a) pools all questions; panel (b) subtracts each task's mean logged values before computing correlations. Zero values receive the same floor as in Section~\ref{sec:exp_procedure}.}
  \label{fig:log_attribute_correlations}
  \vspace{-10pt}
\end{figure}

\paragraph{Attribute correlations.}
Figure~\ref{fig:log_attribute_correlations} shows that logged context length is not positively correlated with any of the three properties: it has weak negative correlations with scope ($r=-0.24$) and dispersion ($r=-0.18$), and almost no correlation with reasoning depth ($r=-0.01$). After task-mean centring, these correlations remain non-positive ($-0.16$, $-0.02$ and $-0.01$, respectively). Scope and dispersion have the strongest association ($r=0.55$ overall; $r=0.45$ after centring). These are descriptive relationships between properties, not associations with model scores.

% \section{Performance distribution}
% \label{appendix:performance_distribution}

% \begin{figure}
%     \centering
%     \includegraphics[width=1\linewidth]{figs/model_leaderboard.pdf}
%     \caption{Caption}
%     \label{fig:performance_distribution}
% \end{figure}

\clearpage
\section{Additional Experimental Results}
\label{appendix:additional_results}

\subsection{GLM Fit and Predictive Performance}
\label{appendix:glm_comparison}

Table~\ref{tab:glm_comparison} compares eight fractional-logit specifications, from a constant predictor to the full additive GLM (Equation~\ref{eq:glm}), using five-fold cross-validation on all 6,550 questions. The folds are stratified by task and shared across specifications, keeping all four LLM scores for each question together. Within each fold, log floors and standardisation are computed from training questions only. Cross-entropy compares predictions with each question--LLM pair's observed mean score, weighting questions equally. These are descriptive prediction checks within the represented tasks and LLMs.

\begin{table}[H]
  \caption{Cross-validated binary cross-entropy with fractional score targets and natural logarithms (lower is better). Loss reduction is relative to the constant training-mean predictor. Every GLM includes an intercept; length and the three properties (scope, dispersion and depth) are logged.}
  \label{tab:glm_comparison}
  \centering
  \small
  \begin{tabular}{lrr}
    \toprule
    Predictors & Cross-entropy & Loss reduction (\%) \\
    \midrule
    Constant (training mean) & 0.4196 & 0.00 \\
    Task only & 0.4006 & 4.51 \\
    LLM only & 0.4145 & 1.21 \\
    Length only & 0.4171 & 0.60 \\
    Length + three properties only & 0.4165 & 0.74 \\
    \midrule
    Task + LLM & 0.3954 & 5.77 \\
    Task + LLM + length & 0.3951 & 5.83 \\
    Task + LLM + length + three properties & 0.3928 & 6.38 \\
    \bottomrule
  \end{tabular}
\end{table}

Task identity gives the largest improvement among the standalone predictor sets. The full GLM reduces cross-entropy by 6.38\% relative to the constant predictor and by 0.58\% relative to task, LLM and length alone. The properties therefore improve prediction beyond these controls, although the incremental gain is modest. These percentages describe reductions in predictive loss, not classification accuracy or variance explained.

\begin{figure}[t]
  \centering
  \includegraphics[width=0.9\linewidth]{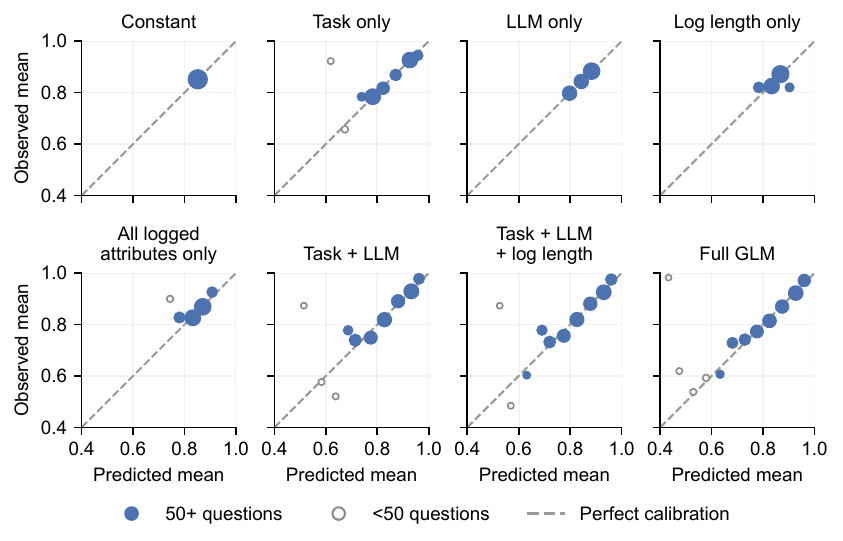}
  \caption{Out-of-fold calibration for the eight specifications in Table~\ref{tab:glm_comparison}. Points show mean predictions and observed scores in prediction bins of width 0.05; the dashed line denotes equality. Larger markers indicate more question--LLM pairs, and hollow markers indicate fewer than 50 distinct questions. All nonempty bins are shown.}
  \label{fig:glm_calibration}
\end{figure}

Figure~\ref{fig:glm_calibration} compares predicted and observed mean scores. The constant predictor is approximately calibrated overall but gives nearly the same prediction to every question--LLM pair. The full GLM broadly follows the diagonal in well-populated bins, although deviations remain: predictions averaging 0.68 have an observed mean of 0.73. Calibration complements cross-entropy by showing where predicted means agree with observed averages.

\subsection{RQ1: Attribute Contrasts}
\label{appendix:rq1}

\begin{table}[H]
  \caption{Supplementary RQ1 score contrasts (pp; 95\% CIs). Top: interquartile contrasts on the same 3,471 questions in the nine tasks with distinct depth quartiles, using the full-data GLM. Bottom: doubling contrasts, which average the predicted score change when a predictor is doubled, over questions with positive values whose doubles remain within their task's observed range. Coverage: context length 34\%, evidence scope 91\%, evidence dispersion 49\%, reasoning depth 60\% of questions. Per-LLM rows use separate fits.}
  \label{tab:rq1_appendix}
  \begin{center}
    \small
    \setlength{\tabcolsep}{3pt}
    \resizebox{\linewidth}{!}{%
      \begin{tabular}{l r r r r}
        \toprule
        & Context length & Evidence scope & Evidence dispersion & Reasoning depth \\
        \midrule
        Common tasks, Q25 $\to$ Q75 & $-2.12\ [-3.19, -1.06]$ & $-2.72\ [-3.90, -1.54]$ & $-0.77\ [-1.28, -0.25]$ & $-3.02\ [-4.44, -1.61]$ \\
        \midrule
        All LLMs, doubling & $-2.13\ [-3.22, -1.05]$ & $-1.04\ [-1.51, -0.57]$ & $-0.28\ [-0.47, -0.09]$ & $-2.70\ [-4.04, -1.36]$ \\
        Gemini 3.8 Flash, doubling & $-0.20\ [-1.38, \phantom{-}0.98]$ & $0.74\ [\phantom{-}0.21, \phantom{-}1.26]$ & $-0.58\ [-0.85, -0.32]$ & $-2.13\ [-3.71, -0.55]$ \\
        GPT-5.6 Luna, doubling & $-1.78\ [-3.08, -0.48]$ & $-1.53\ [-2.08, -0.98]$ & $-0.15\ [-0.38, \phantom{-}0.07]$ & $-2.72\ [-4.27, -1.17]$ \\
        Qwen3.8-27B, doubling & $-1.40\ [-2.44, -0.36]$ & $-0.67\ [-1.16, -0.18]$ & $-0.31\ [-0.54, -0.08]$ & $-0.88\ [-2.06, \phantom{-}0.30]$ \\
        Qwen3.5-4B, doubling & $-5.14\ [-6.76, -3.52]$ & $-2.69\ [-3.33, -2.04]$ & $-0.17\ [-0.41, \phantom{-}0.07]$ & $-5.23\ [-7.11, -3.35]$ \\
        \bottomrule
    \end{tabular}}
  \end{center}
\end{table}

\subsection{RQ2: Interaction and Sensitivity Checks}
\label{appendix:rq2}

\paragraph{Score-scale differences and the logistic link.}
The nonlinear logistic link allows the additive GLM to produce different length contrasts at low and high attribute values despite a constant logit-scale length slope. Table~\ref{tab:rq2_interactions} reports this \emph{link-only} baseline using the same scenarios as Table~\ref{tab:rq2}, alongside the interaction coefficients. The baseline and interaction estimates come from separately fitted models.

\begin{table}[H]
  \caption{RQ2 additive-GLM differences between length contrasts (\emph{link only}, pp) and logit interaction coefficients $\hat\gamma_{Lj}$ from Equation~\ref{eq:glm_gamma}. Brackets give 95\% CIs; $p_{\mathrm{adj}}$ tests $\gamma_{Lj}=0$ using the same six-test Holm family as Table~\ref{tab:rq2}.}
  \label{tab:rq2_interactions}
  \begin{center}
    \small
    \setlength{\tabcolsep}{4pt}
    \begin{tabular}{l r r r}
      \toprule
      Attribute & Link only (pp) & $\hat\gamma_{Lj}$ & $p_{\mathrm{adj}}$ \\
      \midrule
      Evidence scope & $-0.16\ [-0.28, -0.04]$ & $-0.070\ [-0.121, -0.019]$ & $2.1\times10^{-2}$ \\
      Evidence dispersion & $-0.10\ [-0.18, -0.02]$ & $-0.059\ [-0.101, -0.018]$ & $2.1\times10^{-2}$ \\
      Reasoning depth & $-0.13\ [-0.23, -0.04]$ & $-0.028\ [-0.068, \phantom{-}0.012]$ & $1.8\times10^{-1}$ \\
      \bottomrule
    \end{tabular}
  \end{center}
\end{table}

\paragraph{Averaging population.}
Using the same fitted model, restricting the depth comparison to the nine tasks with distinct depth quartiles (3{,}471 questions, 53\%) gives a difference of $-1.25$\,pp (Table~\ref{tab:rq2_robustness}). The estimate and standard error scale by the same factor, leaving the test unchanged. Scope and dispersion have distinct quartiles in all 17 tasks.

\paragraph{Alternative contrasts and attribute definitions.}
Table~\ref{tab:rq2_robustness} reports descriptive sensitivity checks. Doubling length uses the same range rule as Table~\ref{tab:rq1_appendix}. It also keeps only tasks with distinct attribute quartiles in which each of the four cells formed by splitting length and the attribute at their within-task medians has at least five questions. This retains 34\% of questions for scope and dispersion and 8\% for depth. The evidence-token check replaces relative scope with the number of evidence tokens in both the model and the contrast.

The model-free check splits questions at their within-task length and attribute medians, assigning ties to the lower group. It computes the longer-minus-shorter mean score difference within each attribute group, averages across tasks weighted by the number of questions in that group, then subtracts the low-attribute result from the high-attribute result. Tasks must contain all four cells, retaining all questions for scope and dispersion and 93\% for depth. Scores are averaged across LLMs per question; intervals use 1{,}000 bootstrap resamples of questions within task.

\paragraph{Variation across LLMs.}
Per-LLM estimates come from separate fits using the same all-question averaging as the primary contrasts. For Gemini~3.8 Flash, the length contrast is positive at low scope ($+1.21$\,pp) and low dispersion ($+0.64$\,pp), compared with $-0.04$ and $-0.85$\,pp at their high values.

\begin{table}[H]
  \caption{RQ2 sensitivity checks and per-LLM differences between length contrasts (high minus low, pp). Brackets give individual, unadjusted 95\% CIs. These checks are descriptive.}
  \label{tab:rq2_robustness}
  \begin{center}
    \small
    \setlength{\tabcolsep}{3pt}
    \resizebox{\linewidth}{!}{%
      \begin{tabular}{l r r r}
        \toprule
        Check & Evidence scope & Evidence dispersion & Reasoning depth \\
        \midrule
        Distinct depth quartiles & -- & -- & $-1.25\ [-2.70, \phantom{-}0.19]$ \\
        Doubling length & $-1.29\ [-2.03, -0.55]$ & $-1.47\ [-2.39, -0.54]$ & $-1.73\ [-3.73, \phantom{-}0.27]$ \\
        Evidence tokens & $-0.99\ [-1.50, -0.47]$ & -- & -- \\
        Within-task median splits & $-1.58\ [-3.94, \phantom{-}0.86]$ & $-2.21\ [-4.52, \phantom{-}0.09]$ & $-3.60\ [-7.54, \phantom{-}0.16]$ \\
        \midrule
        Gemini 3.8 Flash & $-1.25\ [-1.92, -0.59]$ & $-1.50\ [-2.32, -0.67]$ & $-0.33\ [-1.53, \phantom{-}0.86]$ \\
        GPT-5.6 Luna & $-0.98\ [-1.64, -0.32]$ & $-1.00\ [-1.73, -0.27]$ & $-0.34\ [-1.20, \phantom{-}0.51]$ \\
        Qwen3.8-27B & $-0.75\ [-1.34, -0.17]$ & $-0.50\ [-1.20, \phantom{-}0.20]$ & $-0.99\ [-2.09, \phantom{-}0.10]$ \\
        Qwen3.5-4B & $-0.92\ [-1.69, -0.15]$ & $-0.49\ [-1.36, \phantom{-}0.38]$ & $-1.09\ [-2.14, -0.04]$ \\
        \bottomrule
    \end{tabular}}
  \end{center}
\end{table}

\clearpage
\subsection{RQ3: Exploratory Configurations}
\label{appendix:rq3}

\paragraph{Estimation and uncertainty.}
Out-of-fold predictions use the five folds of Appendix~\ref{appendix:glm_comparison}. Cell and configuration intervals use standard errors across question-level averages, treating predictions and centring values as fixed; they provide individual, not simultaneous, coverage.

\paragraph{Three-attribute configurations.}
Table~\ref{tab:rq3_configurations} groups questions by how many properties exceed their within-task median and whether any exceeds its within-task 90th percentile (\emph{extreme}); ties count as not exceeding the threshold. A separate GLM predicts whether the score is non-zero; zero means all three responses scored zero.
For each LLM, we also add an all-three-high indicator to Equation~\ref{eq:glm}. The bottom rows contrast predictions with this indicator set to one versus zero for the 357 qualifying questions. These fits use the full data and question-clustered CIs.

\begin{table}[H]
  \caption{RQ3 configuration estimates. Observed and predicted outcomes are percentages; centred outcomes, residuals, and indicator contrasts are in pp. Brackets give individual, unadjusted 95\% CIs. Definitions and estimation details are given above.}
  \label{tab:rq3_configurations}
  \begin{center}
    \small
    \setlength{\tabcolsep}{3pt}
    \resizebox{\linewidth}{!}{%
      \begin{tabular}{c c r r r r r}
        \toprule
        Any extreme & \# above median & Questions & Observed & Centred & Predicted & Residual \\
        \midrule
        \multicolumn{7}{l}{\emph{Mean score (\%)}} \\
        no & 0 & 2,212 & $87.9$ & $2.2\ [\phantom{-}1.3, \phantom{-}3.2]$ & $87.8$ & $0.1\ [-0.8, \phantom{-}1.1]$ \\
        no & 1 & 1,504 & $84.6$ & $0.6\ [-0.5, \phantom{-}1.8]$ & $84.2$ & $0.4\ [-0.8, \phantom{-}1.6]$ \\
        no & 2 & 1,328 & $86.2$ & $0.1\ [-1.0, \phantom{-}1.3]$ & $84.9$ & $1.3\ [\phantom{-}0.1, \phantom{-}2.5]$ \\
        no & 3 & 164 & $76.1$ & $-5.4\ [-9.6, -1.3]$ & $77.7$ & $-1.5\ [-5.7, \phantom{-}2.6]$ \\
        yes & 1 & 345 & $81.5$ & $-2.8\ [-5.4, -0.2]$ & $83.8$ & $-2.3\ [-4.9, \phantom{-}0.4]$ \\
        yes & 2 & 804 & $82.2$ & $-3.5\ [-5.2, -1.8]$ & $83.6$ & $-1.4\ [-3.1, \phantom{-}0.3]$ \\
        yes & 3 & 193 & $78.2$ & $-7.0\ [-10.8, -3.2]$ & $80.4$ & $-2.2\ [-5.9, \phantom{-}1.5]$ \\
        \midrule
        \multicolumn{7}{l}{\emph{Share of question--LLM pairs with a non-zero score (\%)}} \\
        -- & 0 & 2,212 & $94.4$ & $0.8\ [-0.0, \phantom{-}1.6]$ & $94.8$ & $-0.4\ [-1.2, \phantom{-}0.4]$ \\
        -- & 1 & 1,849 & $94.3$ & $0.5\ [-0.4, \phantom{-}1.4]$ & $93.9$ & $0.4\ [-0.4, \phantom{-}1.3]$ \\
        -- & 2 & 2,132 & $93.5$ & $-0.4\ [-1.2, \phantom{-}0.4]$ & $93.1$ & $0.3\ [-0.5, \phantom{-}1.2]$ \\
        -- & 3 & 357 & $87.3$ & $-5.0\ [-7.7, -2.2]$ & $89.0$ & $-1.7\ [-4.4, \phantom{-}1.0]$ \\
        \midrule
        \multicolumn{7}{l}{\emph{All-three-high indicator, separate model per LLM (pp)}} \\
        \multicolumn{6}{l}{Gemini 3.8 Flash} & $-1.8\ [-5.4, \phantom{-}1.9]$ \\
        \multicolumn{6}{l}{GPT-5.6 Luna} & $-3.6\ [-8.0, \phantom{-}0.8]$ \\
        \multicolumn{6}{l}{Qwen3.8-27B} & $-2.0\ [-6.0, \phantom{-}1.9]$ \\
        \multicolumn{6}{l}{Qwen3.5-4B} & $-4.6\ [-9.4, \phantom{-}0.3]$ \\
        \bottomrule
    \end{tabular}}
  \end{center}
\end{table}

\paragraph{Sensitivity to task composition.}
Table~\ref{tab:rq3_sensitivity} recomputes observed-score contrasts after omitting one task at a time and within SciTrek alone. Bins and task--LLM centring are retained; the GLM is not refitted.

\begin{table}[H]
  \caption{RQ3 sensitivity to task composition (pp). First three rows: high/high minus low/low centred scores. Last two: high minus low depth at the stated dispersion level. The final column is the range when omitting any single task other than SciTrek; it is not a confidence interval.}
  \label{tab:rq3_sensitivity}
  \begin{center}
    \small
    \setlength{\tabcolsep}{3pt}
    \begin{tabular}{l r r r r}
      \toprule
      Comparison & All tasks & No SciTrek & Only SciTrek & Other omissions \\
      \midrule
      Scope $\times$ dispersion & $-6.5$ & $-6.2$ & $-8.6$ & $[-7.2, -6.1]$ \\
      Scope $\times$ depth & $-6.8$ & $-3.4$ & $-10.2$ & $[-7.4, -6.4]$ \\
      Dispersion $\times$ depth & $-10.0$ & $-4.1$ & $-14.7$ & $[-11.4, -9.8]$ \\
      Depth at low dispersion & $1.0$ & $1.1$ & $0.1$ & $[0.1, 3.0]$ \\
      Depth at high dispersion & $-6.7$ & $-0.8$ & $-12.6$ & $[-7.7, -6.7]$ \\
      \bottomrule
    \end{tabular}
  \end{center}
\end{table}

\clearpage
\paragraph{Pairwise cell estimates.}
Table~\ref{tab:rq3_pair_cells} provides the data behind Figure~\ref{fig:rq3_heatmaps}. Tercile thresholds use unique questions within each task. Ties enter the lower bin, so discrete properties can produce unequal or empty bins; cells with fewer than 10 questions are omitted from the figure.

\begin{table}[H]
  \caption{Cell data for Figure~\ref{fig:rq3_heatmaps}. Horizontal and vertical levels follow each panel's axes. Scores are centred on the task--LLM mean; residuals are observed minus out-of-fold additive-GLM predictions. Brackets give individual, unadjusted 95\% residual CIs (pp).}
  \label{tab:rq3_pair_cells}
  \begin{center}
    \small
    \setlength{\tabcolsep}{5pt}
    \begin{tabular}{l l r r r}
      \toprule
      Horizontal & Vertical & Questions & Centred score (pp) & Residual (pp) \\
      \midrule
      \multicolumn{5}{l}{\emph{(a) Evidence scope $\times$ evidence dispersion}} \\
      low & low & 1,323 & $3.6$ & $0.9\ [-0.2, \phantom{-}2.0]$ \\
      low & mid & 627 & $0.9$ & $-0.0\ [-1.9, \phantom{-}1.9]$ \\
      low & high & 295 & $-0.2$ & $0.2\ [-2.6, \phantom{-}2.9]$ \\
      mid & low & 612 & $0.7$ & $-0.7\ [-2.5, \phantom{-}1.2]$ \\
      mid & mid & 792 & $-0.3$ & $0.0\ [-1.6, \phantom{-}1.7]$ \\
      mid & high & 745 & $-1.4$ & $0.0\ [-1.8, \phantom{-}1.8]$ \\
      high & low & 266 & $-2.0$ & $-2.2\ [-5.2, \phantom{-}0.8]$ \\
      high & mid & 770 & $-0.7$ & $0.8\ [-0.8, \phantom{-}2.4]$ \\
      high & high & 1,120 & $-2.9$ & $-0.8\ [-2.2, \phantom{-}0.7]$ \\
      \midrule
      \multicolumn{5}{l}{\emph{(b) Evidence scope $\times$ reasoning depth}} \\
      low & low & 2,064 & $2.3$ & $0.4\ [-0.5, \phantom{-}1.4]$ \\
      low & mid & 68 & $2.6$ & $3.1\ [\phantom{-}0.9, \phantom{-}5.4]$ \\
      low & high & 113 & $2.0$ & $1.3\ [-3.3, \phantom{-}6.0]$ \\
      mid & low & 1,842 & $0.2$ & $0.1\ [-1.0, \phantom{-}1.1]$ \\
      mid & mid & 102 & $-1.1$ & $0.4\ [-2.0, \phantom{-}2.9]$ \\
      mid & high & 205 & $-5.9$ & $-2.7\ [-6.7, \phantom{-}1.4]$ \\
      high & low & 1,766 & $-1.6$ & $-0.5\ [-1.6, \phantom{-}0.6]$ \\
      high & mid & 134 & $-3.0$ & $-0.2\ [-3.0, \phantom{-}2.6]$ \\
      high & high & 256 & $-4.4$ & $0.1\ [-2.9, \phantom{-}3.2]$ \\
      \midrule
      \multicolumn{5}{l}{\emph{(c) Evidence dispersion $\times$ reasoning depth}} \\
      low & low & 2,016 & $2.0$ & $-0.1\ [-1.1, \phantom{-}0.8]$ \\
      low & mid & 62 & $3.3$ & $2.8\ [\phantom{-}0.2, \phantom{-}5.3]$ \\
      low & high & 123 & $3.0$ & $2.8\ [-0.8, \phantom{-}6.3]$ \\
      mid & low & 1,842 & $0.3$ & $0.3\ [-0.8, \phantom{-}1.4]$ \\
      mid & mid & 125 & $-1.7$ & $0.1\ [-2.7, \phantom{-}2.9]$ \\
      mid & high & 222 & $-2.9$ & $0.2\ [-3.3, \phantom{-}3.7]$ \\
      high & low & 1,814 & $-1.2$ & $-0.1\ [-1.2, \phantom{-}1.1]$ \\
      high & mid & 117 & $-2.8$ & $0.4\ [-2.0, \phantom{-}2.8]$ \\
      high & high & 229 & $-8.0$ & $-3.2\ [-7.1, \phantom{-}0.6]$ \\
      \midrule
      \multicolumn{5}{l}{\emph{(d) Evidence dispersion $\times$ reasoning depth, without SciTrek}} \\
      low & low & 1,709 & $1.8$ & $0.1\ [-0.9, \phantom{-}1.1]$ \\
      low & mid & 62 & $3.3$ & $2.8\ [\phantom{-}0.2, \phantom{-}5.3]$ \\
      low & high & 96 & $2.9$ & $3.0\ [-0.5, \phantom{-}6.5]$ \\
      mid & low & 1,639 & $-0.1$ & $-0.1\ [-1.2, \phantom{-}1.0]$ \\
      mid & mid & 125 & $-1.7$ & $0.1\ [-2.7, \phantom{-}2.9]$ \\
      mid & high & 78 & $-1.7$ & $0.5\ [-3.6, \phantom{-}4.7]$ \\
      high & low & 1,637 & $-1.5$ & $-0.4\ [-1.6, \phantom{-}0.7]$ \\
      high & mid & 117 & $-2.8$ & $0.4\ [-2.0, \phantom{-}2.8]$ \\
      high & high & 87 & $-2.3$ & $1.6\ [-2.5, \phantom{-}5.8]$ \\
      \bottomrule
    \end{tabular}
  \end{center}
\end{table}

\end{document}